\documentclass{article}

\PassOptionsToPackage{numbers,compress}{natbib}
 \usepackage[preprint]{neurips_2025}

\usepackage[utf8]{inputenc} 
\usepackage[T1]{fontenc}    
\usepackage{hyperref}       
\usepackage{url}            
\usepackage{booktabs}       
\usepackage{amsfonts}       
\usepackage{nicefrac}       
\usepackage{microtype}      
\usepackage{xcolor}         

\usepackage{amsmath}   
\usepackage{amssymb}   
\usepackage{amsfonts}  

\usepackage{booktabs}  
\usepackage{array}     

\usepackage{hyperref}  
\usepackage{cleveref}  

\usepackage{graphicx}
\usepackage{caption}

\usepackage{multirow}

\title{EAGLE-Pangu: Accelerator-Safe Tree Speculative Decoding on Ascend NPUs}

\author{%
  Chang Han, Yijie Hu, Jingling Liu \\
  College of Computer Science, Chongqing University
}

\begin{document}

\maketitle

\begin{abstract}

Autoregressive decoding remains a primary bottleneck in large language model (LLM) serving, motivating speculative decoding methods that reduce expensive teacher-model invocations by verifying multiple candidate tokens per step. Tree-structured speculation further increases parallelism, but is often brittle when ported across heterogeneous backends and accelerator stacks, where attention masking, KV-cache layouts, and indexing semantics are not interchangeable.

We present \textsc{EAGLE-Pangu}, a reproducible system that ports EAGLE-3-style tree speculative decoding to a Pangu teacher backend on Ascend NPUs. \textsc{EAGLE-Pangu} contributes (i) an explicit branch/commit cache manager built on the \texttt{Cache} API, (ii) accelerator-safe tree tensorization that removes undefined negative indices by construction and validates structural invariants, and (iii) a fused-kernel-compatible teacher verification path with a debuggable eager fallback.

On 240 turns from MT-Bench and HumanEval-style prompts, \textsc{EAGLE-Pangu} improves end-to-end decoding throughput by \(1.27\times\) on average (up to \(2.46\times\) at p99) over teacher-only greedy decoding in the fused-kernel performance path. We also provide a fused-kernel-free reference path with structured traces and invariant checks to support reproducible debugging and ablation across execution modes and tree budgets.

\end{abstract}

\section{Introduction}

The dominant cost in LLM serving arises from auto-regressive decoding, where each new token requires an additional forward pass through a large ``teacher'' model. This sequential dependency fundamentally limits throughput and increases latency under realistic deployment constraints (e.g., long contexts, high concurrency, and strict tail-latency budgets). Speculative decoding mitigates this bottleneck by using a smaller ``draft'' model to propose candidate tokens that are then verified by the teacher, thereby reducing the number of teacher steps required for a fixed-length generation \cite{speculative_decoding}. Tree-structured speculative decoding further generalizes this idea by validating multiple candidate continuations per step, increasing parallelism and potentially improving throughput beyond linear draft proposals \cite{eagle3,medusa}.

Despite its conceptual simplicity, tree speculative decoding is often fragile in practice. Porting a tree-decoding implementation across model backends and accelerator stacks is not merely an exercise in ``re-implementing kernels'': (i) the teacher model may expose non-standard KV-cache layouts and specialized attention masking interfaces, especially when supporting tree-mode execution; (ii) fused attention kernels impose stricter requirements on mask shapes, alignment, and boundary conditions than eager implementations; and (iii) indexing and gather semantics can differ across device runtimes, where negative or out-of-bounds indices may be unsupported, inconsistently handled, or silently miscomputed. In tree decoding, such discrepancies are amplified because candidate evaluation relies on structured indexing (e.g., retrieving node states along many speculative paths) and on precise masking to prevent cross-branch information leakage. Consequently, a naive port may exhibit degraded quality, sporadic failures, or irreproducible performance, even if it ``runs.''

In this work, we focus on making tree speculative decoding \emph{portable, correct, and reproducible} on a Pangu teacher backend deployed on Ascend NPUs. We follow the EAGLE-3-style tree speculative decoding framework and its posterior acceptance rule \cite{eagle3}, and we do not claim a new decoding algorithm. Instead, we identify the key system-level failure modes that arise in this setting and propose a set of design principles and abstractions that preserve the intended decoding semantics while satisfying accelerator constraints. Our evaluation emphasizes (a) end-to-end decoding throughput, (b) answer quality under matched decoding configurations, and (c) stability and debuggability via structured traces and execution-mode controls.

Concretely, we make the following contributions:
\begin{itemize}
  \item \textbf{Branchable KV-cache abstraction.} We introduce a cache interface that separates accepted-prefix state from per-branch speculative state, enabling correct cache cloning and acceptance updates while decoupling tree decoding from backend-specific KV representations.
  \item \textbf{Accelerator-safe tree tensor semantics.} We propose a correctness-preserving indexing scheme that replaces undefined negative padding indices with a valid dummy index and applies a safe mapping prior to gather operations, complemented by invariant checks to detect shape and boundary violations.
  \item \textbf{Tree-masked teacher execution with fused attention.} We formalize and integrate 4D tree attention masks into the teacher execution path, supporting fused attention kernels for throughput while retaining an eager fallback for verification and debugging.
  \item \textbf{Reproducible distributed pipeline (supporting).} We provide a distributed preprocessing and caching mechanism that avoids redundant work and mitigates synchronization failures in large-scale runs.
  \item \textbf{Data-driven vocabulary subset mapping (supporting).} We implement a reusable draft-vocabulary subset construction and caching workflow to support controlled speed--quality trade-offs in draft modeling.
\end{itemize}

We benchmark \textsc{EAGLE-Pangu} on MT-Bench-style conversational prompts using consistent prompting templates and matched decoding settings for fair comparison. We report tokens-per-second and end-to-end wall-clock speedups relative to standard greedy generation, and we assess quality using the corresponding benchmark scoring protocol. To isolate the impact of each design component, we conduct targeted ablations over (i) cache management strategies, (ii) safe indexing and invariant enforcement, and (iii) fused versus eager attention execution, alongside analyses of tree hyperparameters and draft-vocabulary subset sizes.

The remainder of the paper is organized as follows: Section~\ref{sec:background} reviews speculative decoding and tree verification semantics; Section~\ref{sec:method} presents our cache abstraction, accelerator-safe indexing, and fused teacher verification; Section~\ref{sec:impl} describes the implementation and evaluation harness; Section~\ref{sec:exp} reports experimental results and analyses; finally, Section~\ref{sec:conclusion} concludes and discusses limitations.

\section{Background}
\label{sec:background}

This section reviews speculative decoding and its tree-structured variants, and establishes the notation used throughout the paper. We also highlight the execution-level semantics---KV caching, attention masking, and tensor indexing---that become first-order concerns when porting tree decoding across heterogeneous backends and accelerator stacks.

\subsection{Auto-regressive decoding and KV caching}
\label{sec:background:ar}

Let \(x = (x_1,\ldots,x_T)\) denote a token sequence. An auto-regressive LLM defines a distribution
\[
p_\theta(x) \;=\; \prod_{t=1}^{T} p_\theta(x_t \mid x_{<t}),
\]
and decoding produces tokens sequentially by sampling or selecting
\[
x_t \sim p_\theta(\cdot \mid x_{<t}) \quad \text{or} \quad x_t = \arg\max_{v} p_\theta(v \mid x_{<t}).
\]
The sequential dependency implies that generating \(T\) tokens requires \(T\) teacher forward steps, which typically dominates end-to-end serving latency and constrains throughput.

Modern transformer decoders amortize computation via a key--value (KV) cache. For a prefix \(x_{\le t}\), the cache stores per-layer key/value tensors summarizing attention-relevant history, allowing the model to compute \(p_\theta(\cdot \mid x_{\le t})\) without recomputing attention over all past tokens. We write the teacher transition abstractly as
\[
(\ell_t, \mathcal{C}_t) = \mathrm{TeacherStep}(x_t, \mathcal{C}_{t-1}),
\]
where \(\ell_t \in \mathbb{R}^{|\mathcal{V}|}\) are logits for the next token and \(\mathcal{C}_t\) is the updated KV state.

\subsection{Speculative decoding}
\label{sec:background:spec}

Speculative decoding reduces teacher invocations by pairing a large \emph{teacher} model \(p_\theta\) with a cheaper \emph{draft} model \(q_\phi\) \cite{speculative_decoding}. Informally, the draft proposes multiple candidate tokens, and the teacher verifies them in a batched manner. If the proposed tokens are consistent with the teacher distribution, several tokens may be accepted per teacher step, improving throughput.

Concretely, at a given prefix \(x_{\le t}\), the draft produces a proposal sequence \(y_{1:K} = (y_1,\ldots,y_K)\) using \(q_\phi(\cdot \mid \cdot)\). The teacher then evaluates the proposed continuation in a way that yields teacher probabilities for each position along the proposal. A generic speculative step can be viewed as producing an accepted length \(A \in \{0,\ldots,K\}\) and updating the prefix to
\[
x_{\le t+A} \leftarrow x_{\le t} \,\Vert\, y_{1:A},
\]
where \(\Vert\) denotes concatenation. The specific acceptance rule (e.g., rejection sampling style) is chosen so that the resulting marginal distribution matches (or closely approximates) the teacher distribution under the intended decoding mode \cite{speculative_decoding}.

\subsection{Tree-structured speculative decoding}
\label{sec:background:tree}

Tree-structured speculative decoding generalizes linear proposals to a branching structure, enabling the draft to explore multiple continuations in parallel \cite{eagle3,medusa}. Instead of proposing a single length-\(K\) chain, the draft constructs a rooted tree whose nodes represent candidate next tokens conditioned on their ancestral context.

Let \(\mathcal{T} = (\mathcal{N}, \pi)\) be a rooted tree with node set \(\mathcal{N}\) and parent function \(\pi: \mathcal{N}\setminus\{r\}\to \mathcal{N}\), where \(r\) is the root. Each node \(u \in \mathcal{N}\) (except the root) corresponds to a proposed token \(\tau(u) \in \mathcal{V}\). The depth \(d(u)\) is the number of edges from \(r\) to \(u\). The unique path from \(r\) to \(u\) defines a proposed token sequence
\[
\mathrm{path}(u) = \bigl(\tau(u_1), \tau(u_2), \ldots, \tau(u_{d(u)})\bigr),
\]
where \(u_i\) is the depth-\(i\) ancestor of \(u\). Given a current accepted prefix \(x_{\le t}\), each node corresponds to a candidate continuation \(x_{\le t} \Vert \mathrm{path}(u)\).

The teacher verification step evaluates the tree to obtain teacher logits (or probabilities) for each node under its corresponding ancestral context. We denote this abstractly as
\[
\{\ell(u)\}_{u\in\mathcal{N}} \;=\; \mathrm{TeacherTreeEval}(x_{\le t}, \mathcal{T}, \mathcal{C}_t),
\]
where \(\ell(u)\) provides teacher scores for predicting \(\tau(u)\) at node \(u\), and \(\mathcal{C}_t\) is the KV state of the accepted prefix. A tree acceptance procedure then selects a set of accepted nodes consistent with a valid prefix extension (typically a single root-to-leaf path prefix) and commits the corresponding tokens and cache updates.

Our setting is aligned with EAGLE-3-style tree speculative decoding in which the draft proposes a structured set of candidates and the teacher validates them in a single (or small number of) batched evaluations \cite{eagle3}. We defer the system-specific design choices---especially those required for correctness and portability on our target stack---to Section~\ref{sec:method}.

\subsection{Tree attention masking}
\label{sec:background:mask}

A central requirement for correct tree verification is preventing information leakage across branches. When evaluating multiple speculative nodes in a batched tensor, tokens from different branches must not attend to one another unless they share an ancestor relationship. This is enforced by a \emph{tree attention mask}.

Let the batched speculative tokens be arranged into a tensor of length \(M = |\mathcal{N}\setminus\{r\}|\) (excluding the root), with a fixed node ordering. For two speculative positions corresponding to nodes \(u\) and \(v\), token \(u\) is allowed to attend to token \(v\) if and only if \(v\) lies on the ancestor path of \(u\) (including itself). Denote the ancestor predicate as
\[
\mathrm{Anc}(v,u) =
\begin{cases}
1, & \text{if } v \text{ is an ancestor of } u \text{ (or } v=u\text{)},\\
0, & \text{otherwise.}
\end{cases}
\]
A conceptual attention mask \(M_{\mathrm{tree}}\) can then be defined by assigning \(-\infty\) to disallowed pairs and \(0\) otherwise. In practice, accelerator kernels often expect a broadcastable 4D mask (e.g., \([B, H, M, M]\) or \([B, 1, M, M]\)), and the implementation must preserve the above predicate exactly to maintain semantic correctness.

\subsection{Tensor indexing and gather semantics in tree decoding}
\label{sec:background:index}

Tree decoding relies heavily on structured indexing: mapping each node to its parent, retrieving ancestor-aligned hidden states, and assembling per-path representations for scoring and cache updates. These operations typically compile down to gather/index primitives on the accelerator. Importantly, indexing semantics that are benign in some frameworks (e.g., negative indices as padding sentinels, or implicit wrap-around) may be undefined or unsupported in others, and can lead to silent miscomputations.

A common pattern is to represent parent pointers as an integer array \(\mathrm{parent}[u]\), where the root uses a sentinel value (often \(-1\)) to indicate ``no parent.'' When constructing batched tensors, the sentinel may propagate into gather indices unless explicitly sanitized. On accelerator runtimes where negative indexing is invalid, this can trigger runtime errors; worse, on some stacks it may yield unspecified values. Therefore, correctness requires an indexing scheme that is both semantically faithful to the tree structure and \emph{device-defined} for all possible index values.

\subsection{Problem setting and evaluation perspective}
\label{sec:background:setting}

We consider the deployment setting where the teacher model is served via a Pangu backend and executed on Ascend NPUs. This setting is representative of a broader class of production stacks where (i) attention kernels are heavily fused, (ii) KV caches are stored in backend-specific layouts, and (iii) masking and indexing behaviors must satisfy stricter constraints than in eager GPU implementations. Our objective is to realize tree speculative decoding in a manner that (a) preserves the intended decoding semantics, (b) achieves meaningful throughput improvements, and (c) remains stable and reproducible under distributed execution.

The next section translates these background requirements into concrete system designs: a branchable KV-cache abstraction, accelerator-safe tree tensor semantics for indexing and gather, and a fused-kernel-compatible tree-masked teacher execution path.

\section{Method}
\label{sec:method}

This section presents the core system designs that make tree speculative decoding portable and correct on our target stack. We focus on two building blocks: (i) a \emph{branchable KV-cache} abstraction that cleanly separates committed prefix state from speculative branch state, and (ii) \emph{accelerator-safe tree tensor semantics} that eliminate undefined indexing behaviors while preserving the tree structure required for verification and acceptance. Section~\ref{sec:method:mask} (next) will build on these abstractions to realize fused-kernel-compatible tree-masked teacher execution.

\subsection{Branchable KV-cache abstraction}
\label{sec:method:cache}

Tree speculative decoding stresses KV-cache semantics: a single accepted prefix must support multiple speculative continuations, and the system must be able to (a) evaluate branches without mutating committed state, and (b) efficiently \emph{commit} the cache corresponding to the ultimately accepted tokens. In practice, backend-specific cache layouts and in-place kernel updates make naive cloning brittle: incorrect state sharing can silently pollute branches, whereas full deep copies can erase the throughput benefits of speculative decoding.

We implement a branchable cache manager on top of the HuggingFace \texttt{Cache} interface, maintaining a committed cache \(\mathcal{C}^{\star}\) (\texttt{main\_cache}) and a set of per-candidate branch caches \(\{\mathcal{B}_i\}\) (\texttt{branch\_caches}). At each speculative iteration, we create isolated branch caches by replicating the committed cache state:
\[
\mathcal{B}_i \leftarrow \mathrm{Replicate}(\mathcal{C}^{\star}), \quad i \in \{1,\ldots,N\},
\]
and run teacher verification for different candidate continuations by passing the corresponding \(\mathcal{B}_i\) into the teacher forward path, which extends the cache in-place for that branch. In our current codebase, \(\mathrm{Replicate}(\cdot)\) is implemented via \texttt{deepcopy} for robustness and isolation.

After acceptance, we update the committed cache using one of two implementation-aligned commit modes:
\begin{itemize}
  \item \textbf{Length-based commit.} Given an accepted length \(A\), we update the committed cache by keeping the original committed prefix and adopting only the first \(A\) newly generated steps from the selected branch.
  \item \textbf{Path-index-based commit.} When the tree acceptance stage produces an explicit index mapping \(\mathrm{path\_indices}\), we rebuild the committed cache by reordering the selected branch cache so that the next-step prefix cache matches the accepted path exactly.
\end{itemize}

We enforce the following invariants:
\begin{enumerate}
  \item \textbf{Isolation.} Branch caches are independent replicas of the committed cache. Extending one branch does not mutate \(\mathcal{C}^{\star}\) or other branches.
  \item \textbf{Commit equivalence.} The committed cache after acceptance is equivalent to the cache obtained by sequential teacher execution on the accepted prefix (under the same backend and cache format).
  \item \textbf{Backend transparency (via Cache API).} The implementation relies only on backend-agnostic \texttt{Cache} primitives (e.g., \texttt{get\_seq\_length} and \texttt{to\_legacy\_cache/from\_legacy\_cache}) to support safe cache rebuilding across different KV layouts.
\end{enumerate}

\paragraph{Commit by path indices and prefix-sharing fast reorder.}
In practice, tree decoding frequently accepts only a short suffix while retaining a long committed prefix. A naive path-index-based commit would reorder the entire prefix cache, incurring unnecessary memory movement. We therefore implement two commit paths for \(\mathrm{path\_indices}\):
\begin{itemize}
  \item \textbf{Full reordering (general).} We rebuild the cache by gathering KV states along the sequence dimension according to \(\mathrm{path\_indices}\) using \texttt{to\_legacy\_cache/from\_legacy\_cache}.
  \item \textbf{Prefix-sharing fast reorder (common-case).} When \(\mathrm{path\_indices}\) preserves the already-committed prefix order, we keep the prefix segment as a contiguous slice and only gather the newly accepted segment, then concatenate them to form the new committed cache. This fast path is controlled by an execution flag (e.g., \texttt{EA\_FAST\_CACHE\_REORDER}) and safely falls back to full reordering upon any boundary or shape inconsistency.
\end{itemize}

The cache manager does not construct tree indices by itself. Instead, the tree tensorization and acceptance logic computes \(\mathrm{path\_indices}\), which maps the next-step accepted prefix sequence to positions in the selected branch cache. The cache manager is responsible only for applying this mapping to produce a correct committed cache for the subsequent decoding iteration.


\subsection{Accelerator-safe tree tensor semantics}
\label{sec:method:index}

Tree verification and acceptance require frequent gather operations over node-indexed tensors (e.g., retrieving parent/ancestor-aligned hidden states, assembling per-node path features, and mapping teacher outputs back to node IDs). On some accelerator runtimes, negative indices are undefined, and certain fused kernels require all indices to be statically in-bounds. We therefore design a tree tensorization scheme that (i) preserves the ancestor relationships required by the algorithm, and (ii) guarantees that every index used by device-side gathers is valid by construction.

\paragraph{Node linearization and base arrays.}
Let the speculative nodes (excluding root) be ordered as \(\{u_1,\ldots,u_M\}\), with integer IDs \(\mathrm{id}(u_k)=k\). We store:
\begin{itemize}
  \item \(\mathrm{parent}[k] = \mathrm{id}(\pi(u_k))\), where \(\pi(u_k)\in\{r,u_1,\ldots,u_M\}\);
  \item \(\mathrm{depth}[k] = d(u_k)\);
  \item \(\tau[k] = \tau(u_k)\), the proposed token at node \(u_k\).
\end{itemize}
Many operations can be expressed as gathers on node-indexed tensors \(Z \in \mathbb{R}^{M\times D}\), e.g.,
\[
Z_{\mathrm{par}}[k] = Z[\mathrm{parent}[k]].
\]
The difficulty is that \(\mathrm{parent}[k]\) is not defined for nodes whose parent is the root \(r\).

\paragraph{Dummy-root indexing (sentinel-free gathers).}
We eliminate all sentinel parents by introducing a dummy root row at index \(0\). Concretely, we allocate node-indexed tensors with an extra row, \(\widetilde{Z}\in\mathbb{R}^{(M+1)\times D}\), where row \(0\) represents the root state (the committed prefix). We define a shifted ID map:
\[
\widetilde{\mathrm{id}}(r)=0,\qquad \widetilde{\mathrm{id}}(u_k)=k \ \ (k\in\{1,\ldots,M\}),
\]
and a valid parent array \(\widetilde{\mathrm{parent}}[k]\in\{0,1,\ldots,M\}\) such that
\[
\widetilde{\mathrm{parent}}[k] =
\begin{cases}
0, & \text{if } \pi(u_k)=r,\\
\widetilde{\mathrm{id}}(\pi(u_k)), & \text{otherwise.}
\end{cases}
\]
All gathers are then executed with \(\widetilde{\mathrm{parent}}\), guaranteeing in-range indices on device. The dummy row is populated with the representation corresponding to the current accepted prefix (i.e., the tree root context), ensuring depth-1 nodes correctly reference the prefix state.

\paragraph{Ancestor tables for path-structured operations.}
Beyond immediate parents, tree decoding often needs a bounded number of ancestors (e.g., to construct per-node path features or to build masks). Let \(D_{\max}=\max_k \mathrm{depth}[k]\). We build an ancestor index table \(A\in\mathbb{Z}^{(D_{\max}+1)\times (M+1)}\) with:
\[
A[0,k]=k,\qquad A[\ell+1,k]=\widetilde{\mathrm{parent}}(A[\ell,k]).
\]
By construction, \(A[\ell,k]\in\{0,\ldots,M\}\) for all \(\ell,k\), so subsequent gathers such as \(\widetilde{Z}[A[\ell,k]]\) are accelerator-safe.

\paragraph{Padding and batch semantics.}
For batch size \(B\), each sample \(b\) may have a different number of speculative nodes \(M_b\). We pad to \(M_{\max}=\max_b M_b\), and maintain a boolean validity mask \(\mathrm{valid}_b[k]\) indicating whether node slot \(k\) is active. All arrays (\(\widetilde{\mathrm{parent}}, \mathrm{depth}, \tau\)) are padded with device-defined values (e.g., \(\widetilde{\mathrm{parent}}=0\), \(\mathrm{depth}=0\), \(\tau=\mathrm{pad}\)) and guarded by \(\mathrm{valid}\) in later masking/scoring so that padded slots cannot influence acceptance.

\paragraph{Structural invariants (unit-testable).}
To prevent silent structural corruption (e.g., malformed trees, backend integer overflows), we validate invariants prior to launching fused kernels:
\begin{enumerate}
  \item \textbf{Range.} \(\widetilde{\mathrm{parent}}[k]\in[0,M]\) for all \(k\in[1,M]\).
  \item \textbf{Acyclicity / depth consistency.} For \(k>0\), \(\mathrm{depth}[\widetilde{\mathrm{parent}}[k]] < \mathrm{depth}[k]\), and repeated parent application reaches \(0\) within \(\mathrm{depth}[k]\) steps.
  \item \textbf{Validity closure.} If \(\mathrm{valid}[k]=1\) then \(\mathrm{valid}[\widetilde{\mathrm{parent}}[k]]=1\) (except for the root index \(0\)).
\end{enumerate}
These checks are lightweight relative to a teacher forward and substantially improve debuggability and reproducibility on heterogeneous runtimes.

\paragraph{Complexity.}
The shifted indexing adds one dummy row and replaces sentinel handling with constant-time preprocessing. Building an ancestor table costs \(\mathcal{O}(M D_{\max})\) time and memory. In our setting, \(D_{\max}\) is small (tree depth is bounded by the speculative budget), making this overhead negligible compared to teacher compute.

\subsection{Fused tree-masked teacher execution on Pangu+Ascend}
\label{sec:method:mask}

Given a speculative tree \(\mathcal{T}\) and the committed prefix cache \(\mathcal{C}^{\star}\), the teacher must evaluate all speculative nodes under their correct ancestral contexts without cross-branch leakage. On our target stack, this evaluation must also align with fused attention kernels and backend-specific KV-cache layouts. We implement teacher verification as a single batched forward pass over speculative tokens, with (i) cache reads from the committed prefix, (ii) branch-local cache writes for speculative tokens, and (iii) a tree attention mask that enforces the ancestor predicate.

\paragraph{Execution view: prefix-cache + speculative tokens.}
Let \(t\) be the current accepted prefix length. Conceptually, each speculative node \(u_k\) corresponds to a position \(t + \mathrm{depth}[k]\) along its own root-to-node path. We feed the teacher a tensor of speculative token IDs \(\tau[1{:}M]\) (padded to \(M_{\max}\) under batching) while providing \(\mathcal{C}^{\star}\) as the prefix KV-cache. The teacher computes hidden states for the speculative positions and produces per-node logits \(\ell[k]\) for verifying \(\tau[k]\) under the appropriate context.

\paragraph{Tree attention mask (ancestor-only visibility).}
Within the speculative segment, token \(k\) may attend to token \(j\) if and only if \(j\) is an ancestor of \(k\) (including itself) in the speculative tree (Section~\ref{sec:background:mask}). Using the ancestor table \(A\) from Section~\ref{sec:method:index}, we can express the predicate:
\[
\mathrm{Anc}(j,k)=1 \quad \Leftrightarrow \quad \exists \ell \in [0, D_{\max}] \text{ s.t. } A[\ell,k]=j.
\]
We construct a mask \(M_{\mathrm{tree}}\) over speculative positions such that
\[
M_{\mathrm{tree}}[k,j] =
\begin{cases}
0, & \mathrm{Anc}(j,k)=1 \ \text{and}\ \mathrm{valid}[k]=\mathrm{valid}[j]=1,\\
-\infty, & \text{otherwise.}
\end{cases}
\]
This mask is then combined with the backend-required masks (e.g., padding masks), broadcast to the fused kernel's expected shape (typically \([B,1,M_{\max},M_{\max}]\) or \([B,H,M_{\max},M_{\max}]\)).

\paragraph{No leakage to padded slots.}
For padded node slots, we set \(\mathrm{valid}[k]=0\) and force all attention into/out of \(k\) to be masked. This ensures that arbitrary pad token IDs and dummy indices cannot affect active logits.


\paragraph{KV-cache writes and commit.}
During teacher verification, KV blocks for speculative positions are written into the selected \emph{branch cache} (Section~\ref{sec:method:cache}) as the forward pass extends the cache in-place. After the acceptance step, we update the committed cache either by (i) a length-based commit that keeps only the first \(A\) new steps from the selected branch, or (ii) a path-index-based commit that reorders the selected branch cache to match the accepted path prefix, with a prefix-sharing fast path when the committed prefix order is preserved.

\paragraph{Mapping teacher outputs to node verification scores.}
The fused teacher forward produces logits for the next-token prediction at each speculative slot. We extract, for each node \(k\), the teacher score for the proposed token \(\tau[k]\) under the masked context, yielding \(\ell(k)\) used by the acceptance rule. This extraction is implemented as an in-range gather over the vocabulary dimension and is guarded by \(\mathrm{valid}\) so padded slots are ignored.

\paragraph{Correctness guarantee.}
The construction above enforces two key properties:
\begin{enumerate}
  \item \textbf{Context correctness.} For each node \(k\), the hidden state used to score \(\tau[k]\) depends only on the committed prefix and the speculative tokens on the unique ancestor chain of \(k\).
  \item \textbf{State safety.} Speculative KV writes cannot mutate \(\mathcal{C}^{\star}\); committing updates is an explicit operation performed only after acceptance.
\end{enumerate}
Together with the cache invariants in Section~\ref{sec:method:cache}, committing the accepted path yields the same teacher state as sequentially decoding those accepted tokens from the committed prefix.

\paragraph{Implementation note (dense vs. structured masks).}
A direct dense \(M_{\mathrm{tree}}\in\mathbb{R}^{M_{\max}\times M_{\max}}\) is simplest and matches typical fused-kernel interfaces. When \(M_{\max}\) grows, one may reduce overhead by precomputing compact ancestor encodings (e.g., via \(A\)) and generating the mask on-device; our implementation selects the mask construction strategy based on the speculative budget to balance mask overhead and kernel simplicity.

\section{Implementation \& Evaluation Harness}
\label{sec:impl}

Our primary goal is to make EAGLE-3-style tree speculative decoding \emph{deployable and diagnosable} on the Pangu teacher backend running on Ascend NPUs.
Accordingly, we treat execution-path control and structured tracing as first-class parts of the system, rather than ad-hoc debugging aids.

\subsection{Two-mode execution protocol (reference vs. performance)}
\label{sec:impl:modes}

A recurring theme in our study is that ``tree decoding semantics'' are inseparable from the backend execution path (KV layout, attention kernels, and mask handling).
We therefore adopt an explicit two-mode protocol throughout the implementation and evaluation:

\begin{itemize}
  \item \textbf{Reference mode (fused attention off).}
  Used for isolating semantic bugs and for running invariant checks under a debuggable execution path.
  In this mode we disable fused attention via backend environment flags (e.g., \texttt{PANGU\_DISABLE\_NPU\_FUSED=1} and \texttt{PANGU\_DISABLE\_NPU\_FUSED\_TREE=1}).

  \item \textbf{Performance mode (fused attention on).}
  Used for throughput benchmarks and budget sweeps.
  In this mode fused attention is enabled (e.g., \texttt{PANGU\_DISABLE\_NPU\_FUSED=0} and \texttt{PANGU\_DISABLE\_NPU\_FUSED\_TREE=0}, or unset depending on the runtime default).
\end{itemize}

We also include an \textbf{eager-aligned analysis variant} that forces eager attention (e.g., \texttt{PANGU\_FORCE\_EAGER\_ATTN=1}) and aligns baseline vs.\ EA to follow the same high-level generation path when possible, to support controlled comparisons across execution paths.

\subsection{Vendor override and narrow module boundaries}
\label{sec:impl:vendor}

To keep the system portable and to preserve the production code path, we follow a ``vendor override'' principle: changes required for tracing, profiling, and controlled ablations are implemented in a vendor copy of the relevant modules, and imported by prepending the vendor directory to \texttt{sys.path}.
This keeps the core contracts in Section~\ref{sec:method} stable while allowing per-experiment instrumentation without patching the upstream runtime.

\subsection{Structured traces and debug artifacts}
\label{sec:impl:trace}

Every run produces a \emph{manifest} (hyperparameters, environment flags, and version identifiers) and structured per-prompt/per-turn traces.
The traces capture the key execution- and analysis-facing signals (e.g., decoding configuration, speculative-tree statistics, acceptance summaries, and per-stage timing) to enable reproducible benchmarking and post-hoc diagnosis without relying on ad-hoc logs. When a run terminates abnormally (e.g., runtime error or invariant-check failure), we emit a compact failure dump that records the minimal context needed to reproduce the issue (prompt identifier, inputs, and relevant tree/cache metadata).

\subsection{Distributed evaluation and deterministic sharding}
\label{sec:impl:dist}

We run experiments with multi-process distributed execution (\texttt{torchrun} on 8 NPUs).
To ensure deterministic coverage, prompts are sharded by a stable rule (e.g., \(\mathrm{prompt\_id} \bmod \mathrm{world\_size}\)).
Each rank writes independent trace files, and rank 0 merges them into a globally sorted output to produce summary CSV/JSON tables.

\subsection{Timing methodology on Ascend NPUs}
\label{sec:impl:timing}

Throughput measurements report end-to-end wall-clock time for a full generation call, with device synchronization to avoid asynchronous timing artifacts.
We report:
(i) tokens per second (Tok/s),
(ii) time per output token (TPOT), and
(iii) time to first token (TTFT) when measurable by the runtime.
Profiling experiments (stage breakdown and attention-stat sampling) are run separately because instrumentation perturbs end-to-end timing; we use them only for explanation, not as headline throughput results.

\section{Experiments}
\label{sec:exp}

We report results that address: (i) \textbf{end-to-end throughput} on the fused performance path, (ii) \textbf{budget sensitivity} and practical parameter choices, and (iii) \textbf{explanatory analyses} (overhead breakdown, negative results, and attention evidence).

\subsection{Setup}
\label{sec:setup}

\paragraph{Datasets and sample size.}
We evaluate on 160 prompts (80 HumanEval-style prompts and 80 MT-Bench prompts) totaling 240 turns (MT-Bench has 2 turns).
Prompt and output length statistics are computed from the structured traces (mean prompt length \(\approx 501\) tokens; mean output length \(\approx 891\) tokens; outputs are capped by \texttt{max\_new\_tokens} in each experiment).

\begin{figure}[t]
  \centering
  \begin{minipage}{0.49\linewidth}
    \centering
    \includegraphics[width=\linewidth]{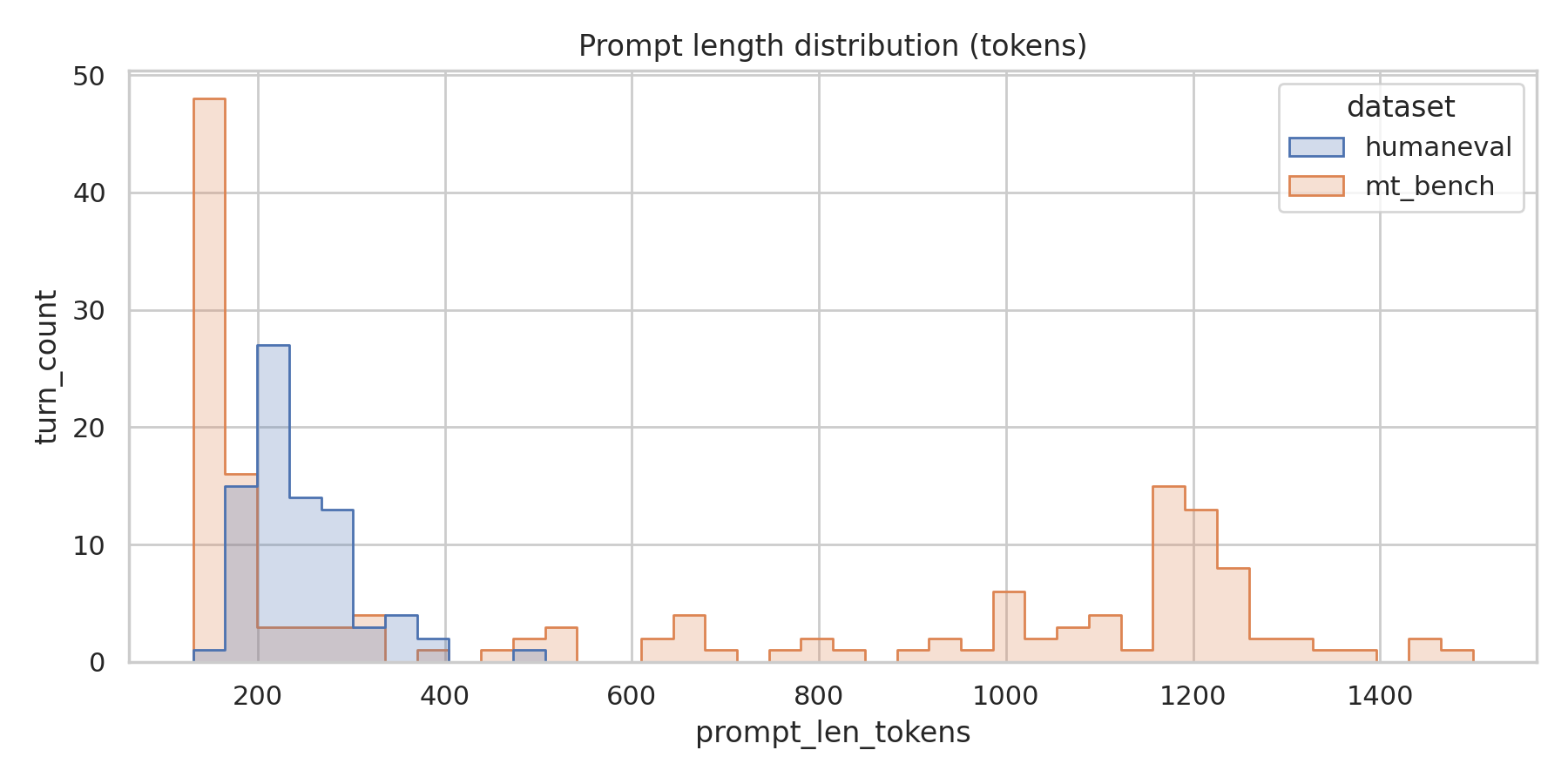}
    \caption*{Prompt length distribution.}
  \end{minipage}\hfill
  \begin{minipage}{0.49\linewidth}
    \centering
    \includegraphics[width=\linewidth]{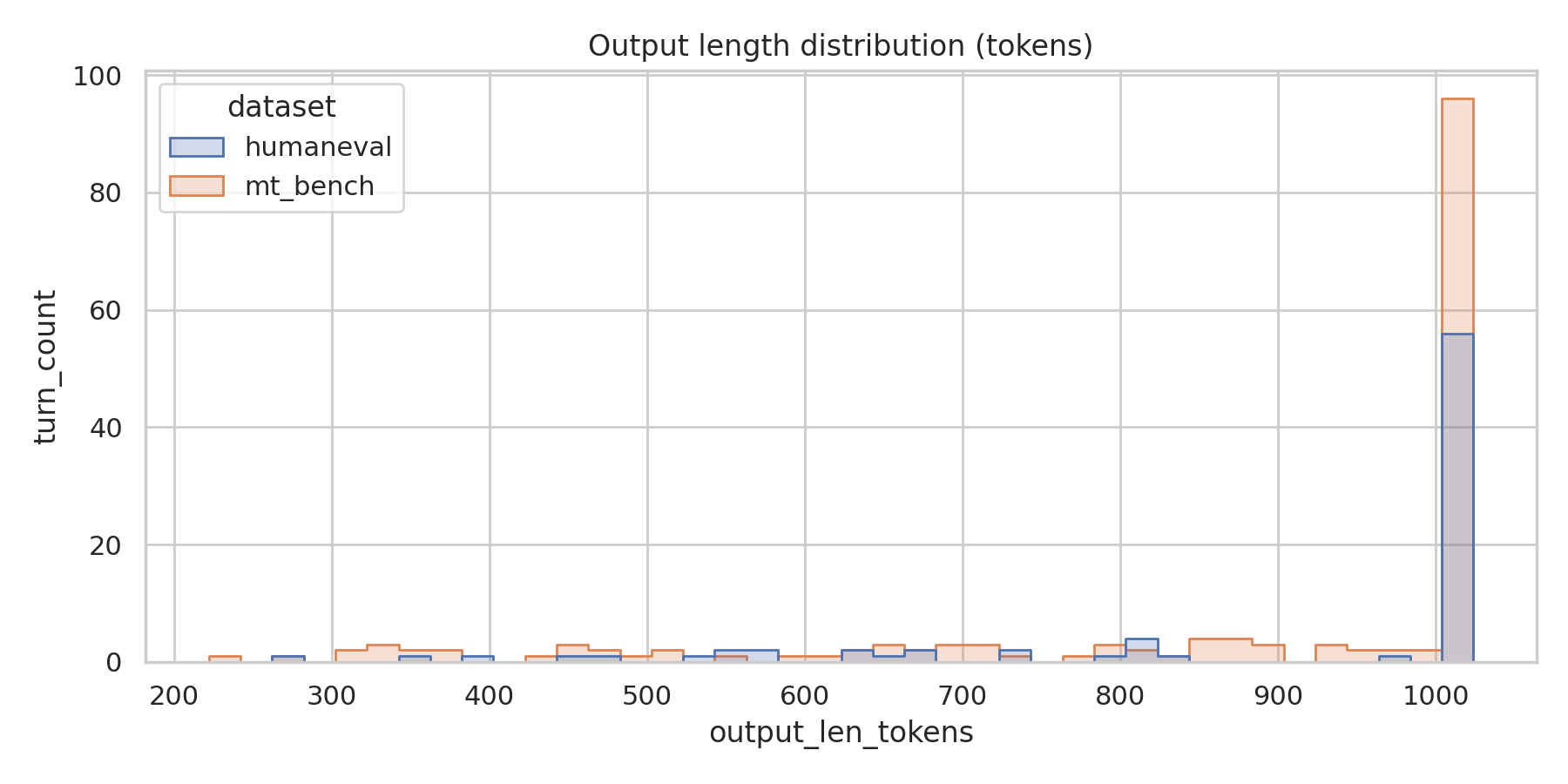}
    \caption*{Output length distribution.}
  \end{minipage}
  \caption{Input/output length distributions in our evaluation set (from structured traces).}
  \label{fig:length_overview}
\end{figure}

\paragraph{Teacher backend and hardware.}
We run the teacher via a Pangu backend on Ascend NPUs, following the deployment setting described in Pangu Embedded \cite{pangu}.

\paragraph{Decoding configuration.}
Unless otherwise noted, we use greedy decoding (\texttt{temperature=0}) and \texttt{max\_new\_tokens=1024}.
Budget sweeps use \texttt{max\_new\_tokens=256} for faster scanning.
We compare:
\textbf{baseline} (teacher-only greedy) vs.\ \textbf{EA} (tree speculative decoding using the same teacher).

\paragraph{Metrics.}
We report Tok/s and speedup (\(\mathrm{Tok/s}_{\mathrm{EA}} / \mathrm{Tok/s}_{\mathrm{base}}\)).
To explain speedups, we report the accepted draft length per verification step, \(L_k\), summarized as \texttt{accept\_L} (flattened across all steps and turns), and the position-wise acceptance curve (\texttt{accept\_pos}).

\subsection{Main results and analyses}
\label{sec:results}

\paragraph{Summary.}
Across 240 turns, we find that tree speculation yields consistent end-to-end throughput gains and the best configuration in our budget sweep reaches a \(1.48\times\) mean speedup, and naive drafter context truncation substantially reduces acceptance and can negate speedups.

\paragraph{E1: End-to-end throughput (batch size 1, fused on).}
\textbf{Goal.} Measure end-to-end decoding throughput improvements in the fused-kernel performance path.
\textbf{Setup.} We run batch-1 generation over 240 turns and report per-turn Tok/s and speedup over teacher-only greedy decoding.

\begin{table}[t]
\centering
\caption{Throughput microbenchmark (240 turns, fused attention enabled). Tok/s and speedup are summarized across turns; \texttt{accept\_L} summarizes flattened \(L_k\) samples across all EA verification steps.}
\label{tab:throughput_micro}
\begin{tabular}{lrrrr}
\toprule
Metric & mean & p50 & p90 & p99 \\
\midrule
Baseline Tok/s & 17.65 & 17.72 & 18.69 & 19.37 \\
EA Tok/s & 22.42 & 21.72 & 31.65 & 42.09 \\
Speedup (\(\times\)) & 1.27 & 1.21 & 1.84 & 2.46 \\
accept\_L (\(L_k\)) & 3.17 & 3 & 6 & 8 \\
\bottomrule
\end{tabular}
\end{table}

\begin{figure}[t]
  \centering
  \begin{minipage}{0.49\linewidth}
    \centering
    \includegraphics[width=\linewidth]{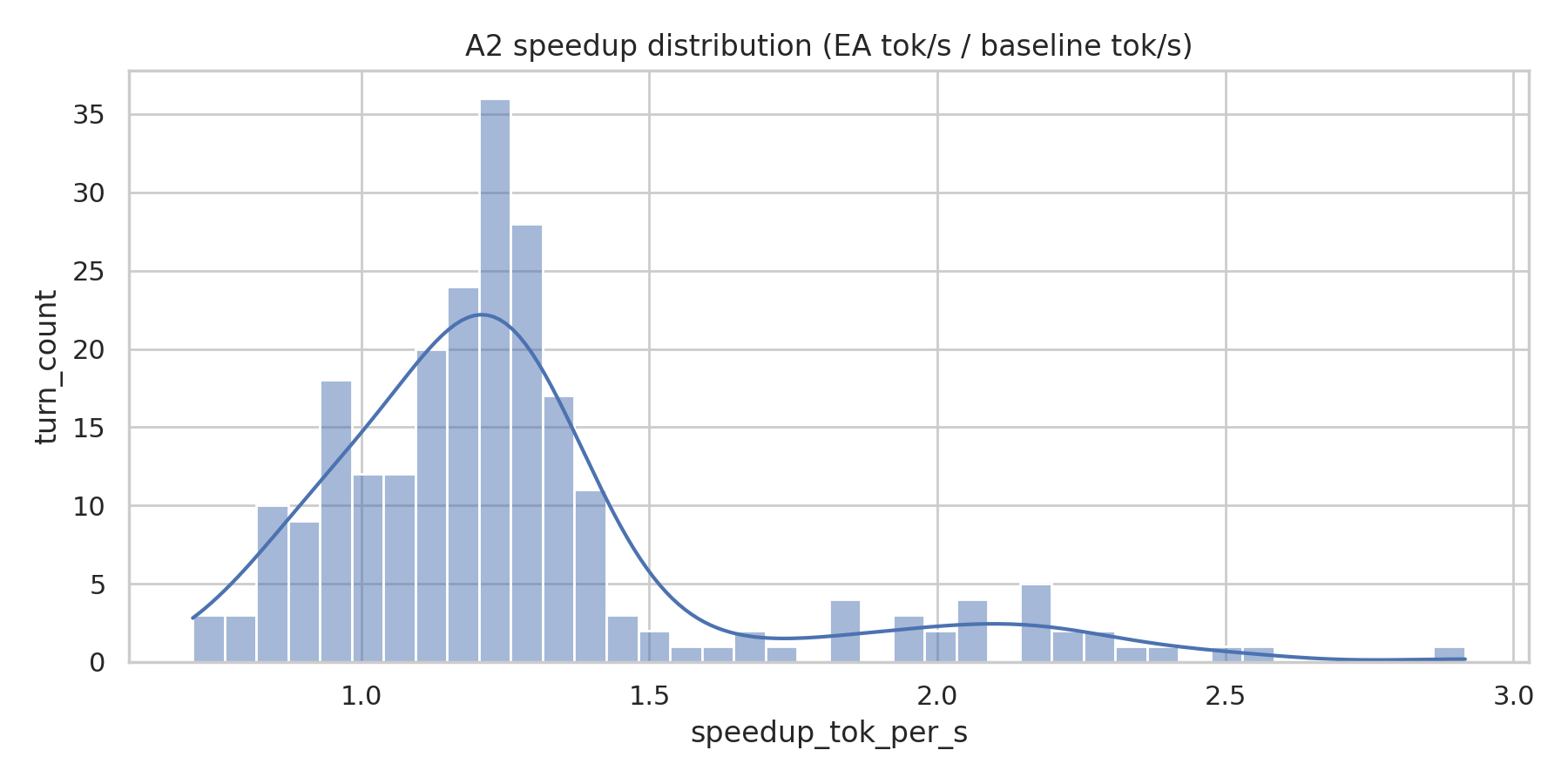}
    \caption*{(a) Speedup distribution.}
  \end{minipage}\hfill
  \begin{minipage}{0.49\linewidth}
    \centering
    \includegraphics[width=\linewidth]{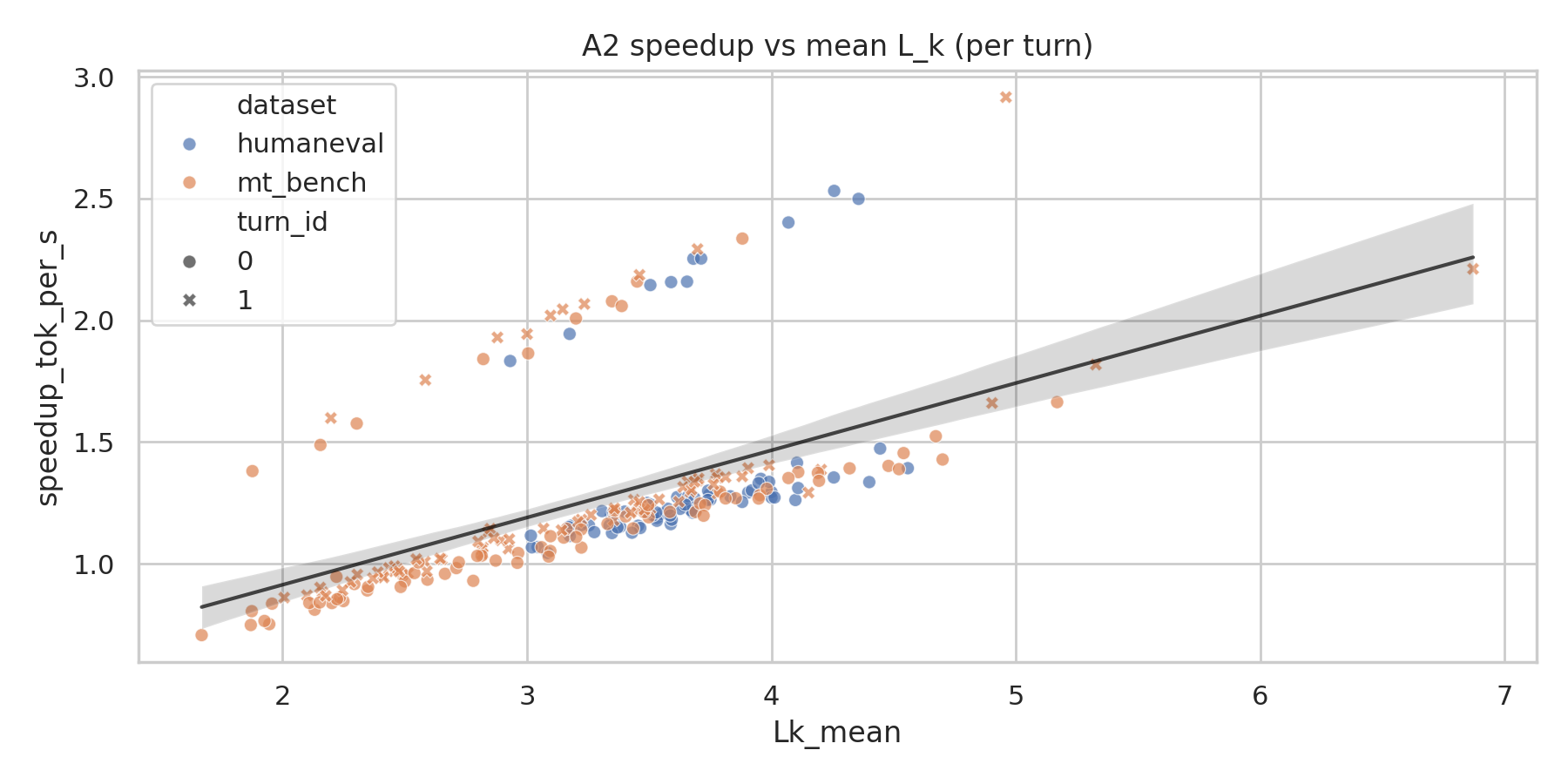}
    \caption*{(b) Speedup vs.\ mean \(L_k\).}
  \end{minipage}
  \caption{Throughput microbenchmark diagnostics (fused on). Speedup correlates with accepted draft length.}
  \label{fig:a2_speedup}
\end{figure}

\begin{figure}[t]
  \centering
  \includegraphics[width=0.78\linewidth]{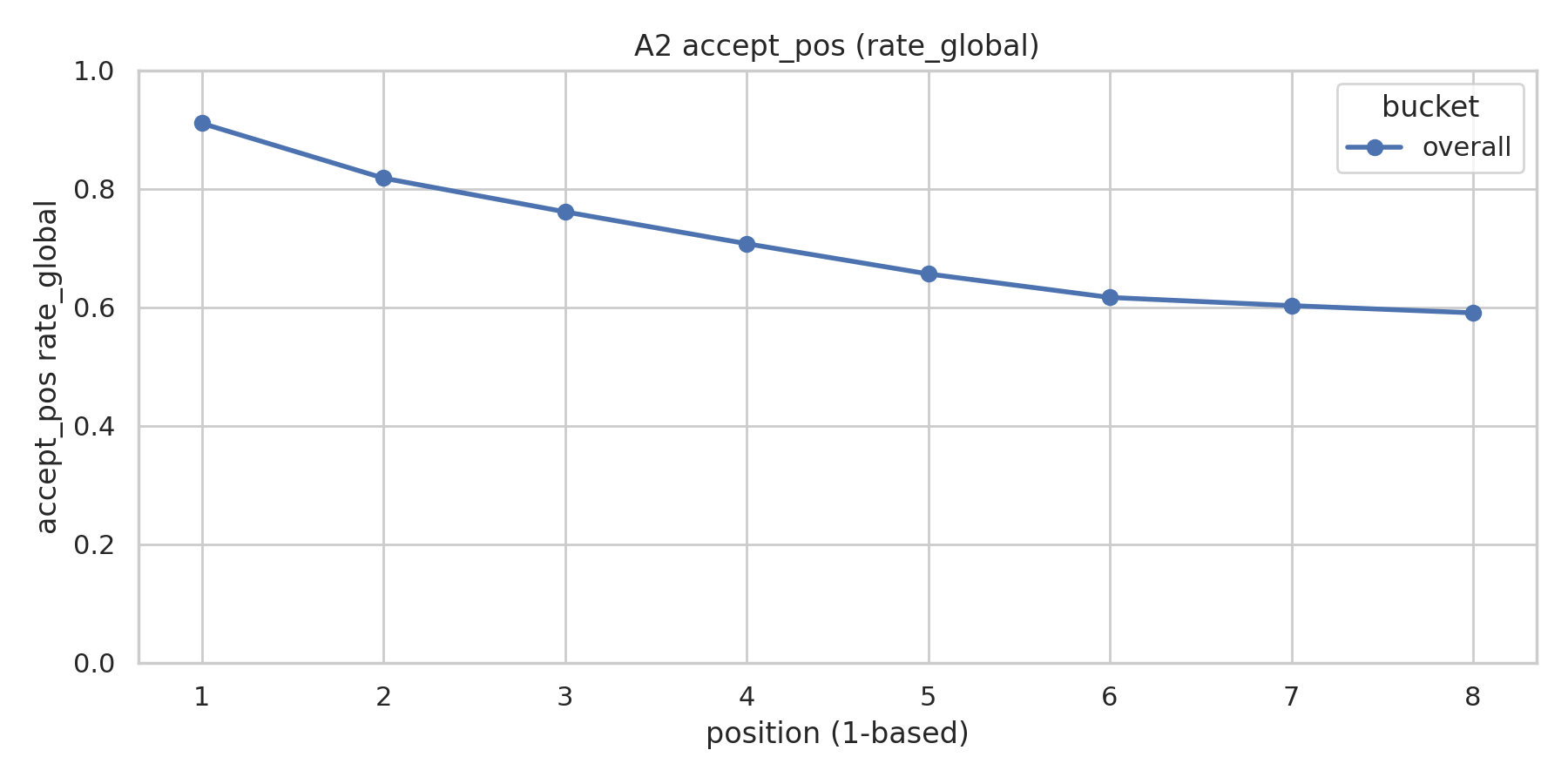}
  \caption{Position-wise acceptance (\texttt{accept\_pos}) aggregated over the evaluation set. Later draft positions are harder to accept, explaining the long-tail behavior of \(L_k\).}
  \label{fig:a2_accept_pos}
\end{figure}


\noindent\textbf{Result.}
\textsc{EAGLE-Pangu} increases mean throughput from 17.65 to 22.42 Tok/s, corresponding to a \(1.27\times\) mean speedup; gains are larger in the tail (p90: \(1.84\times\), p99: \(2.46\times\); Table~\ref{tab:throughput_micro}). The accepted length per verification step has mean 3.17 (median 3; Table~\ref{tab:throughput_micro}) and is positively correlated with speedup (Figure~\ref{fig:a2_speedup}). Position-wise acceptance decays with draft position (Figure~\ref{fig:a2_accept_pos}), explaining diminishing returns for deeper draft positions.

\noindent\textbf{Takeaway.}
Even with a conservative cache-replication implementation, tree speculation yields a clear throughput improvement at batch size 1; acceptance length is the primary driver of per-turn speedup variance.

\paragraph{E2: Budget sensitivity and practical sweet spots (humaneval-only sweep).}
Tree budgets are not ``bigger is better'' due to mask/tensor overheads and diminishing acceptance returns.
We sweep (i) node budget \(M\) with fixed depth bound and (ii) depth bound with fixed \(M\), using \texttt{max\_new\_tokens=256} for efficiency.
\textbf{Goal.} Identify practical tree-budget settings under realistic overheads (masking, tensorization, and commit).
\textbf{Setup.} On the HumanEval subset with \texttt{max\_new\_tokens=256}, we sweep node budget \(M\) (fixed \(D_{\max}=10\)) and depth bound \(D_{\max}\) (fixed \(M=64\)).

\begin{table}[t]
\centering
\caption{Budget sweep summary (humaneval-only, \texttt{max\_new\_tokens=256}, fused on). We report mean Tok/s and mean speedup.}
\label{tab:budget_sweep}
\begin{tabular}{lrrr}
\toprule
Sweep & Setting & EA Tok/s (mean) & Speedup (mean) \\
\midrule
\multirow{5}{*}{Scan \(M\) (fixed \(D_{\max}=10\))} 
  & \(M=16\)  & 27.35 & 1.48 \\
  & \(M=32\)  & 19.54 & 1.06 \\
  & \(M=64\)  & 21.44 & 1.17 \\
  & \(M=128\) & 22.74 & 1.24 \\
  & \(M=256\) & 22.95 & 1.25 \\
\midrule
\multirow{5}{*}{Scan \(D_{\max}\) (fixed \(M=64\))} 
  & \(D_{\max}=4\)  & 19.58 & 1.06 \\
  & \(D_{\max}=8\)  & 22.33 & 1.21 \\
  & \(D_{\max}=10\) & 21.60 & 1.17 \\
  & \(D_{\max}=12\) & 20.52 & 1.11 \\
  & \(D_{\max}=16\) & 18.78 & 1.02 \\
\bottomrule
\end{tabular}
\end{table}


\begin{figure}[t]
  \centering
  \begin{minipage}{0.49\linewidth}
    \centering
    \includegraphics[width=\linewidth]{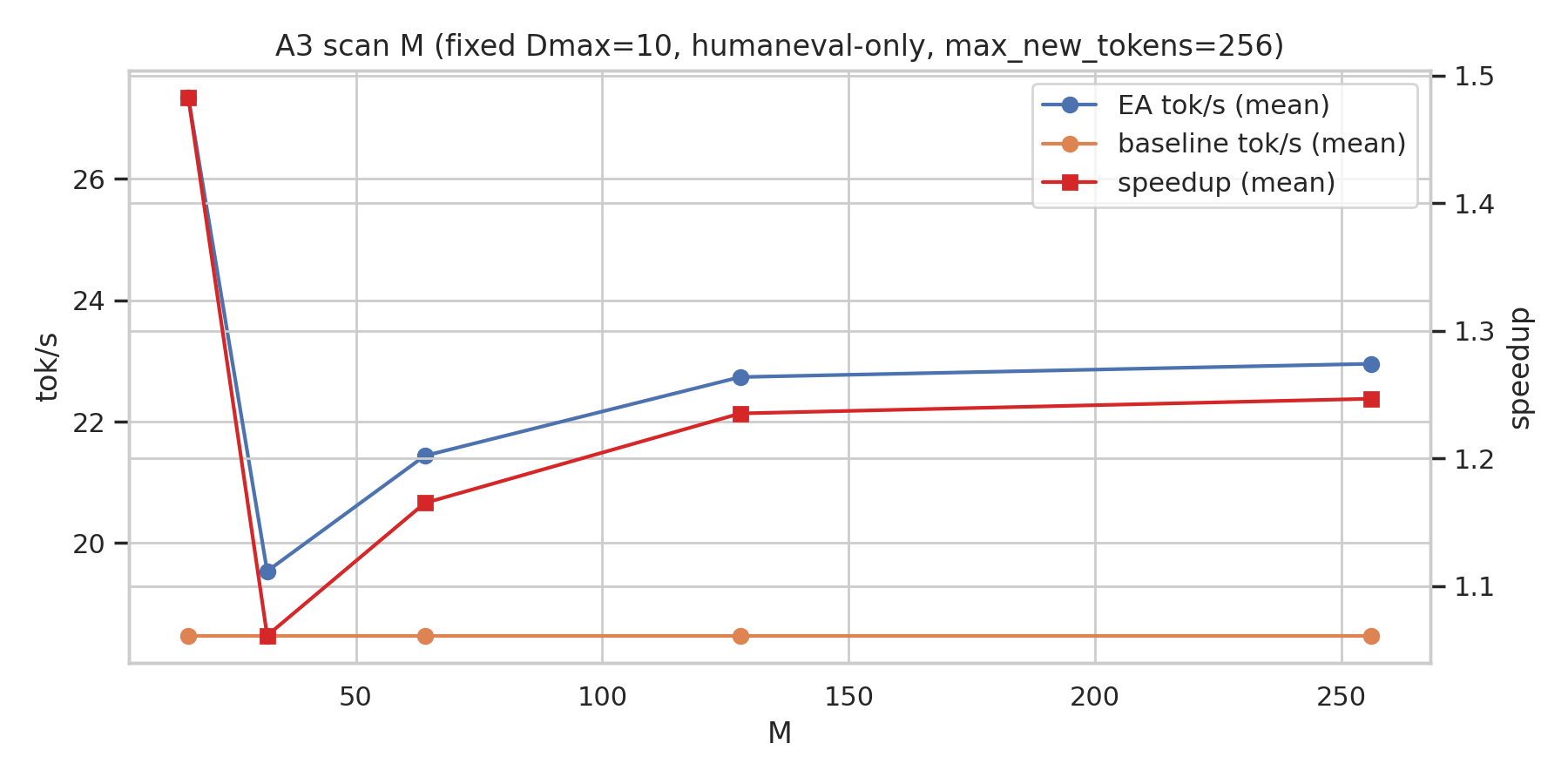}
    \caption*{(a) Scan \(M\) (fixed \(D_{\max}=10\)).}
  \end{minipage}\hfill
  \begin{minipage}{0.49\linewidth}
    \centering
    \includegraphics[width=\linewidth]{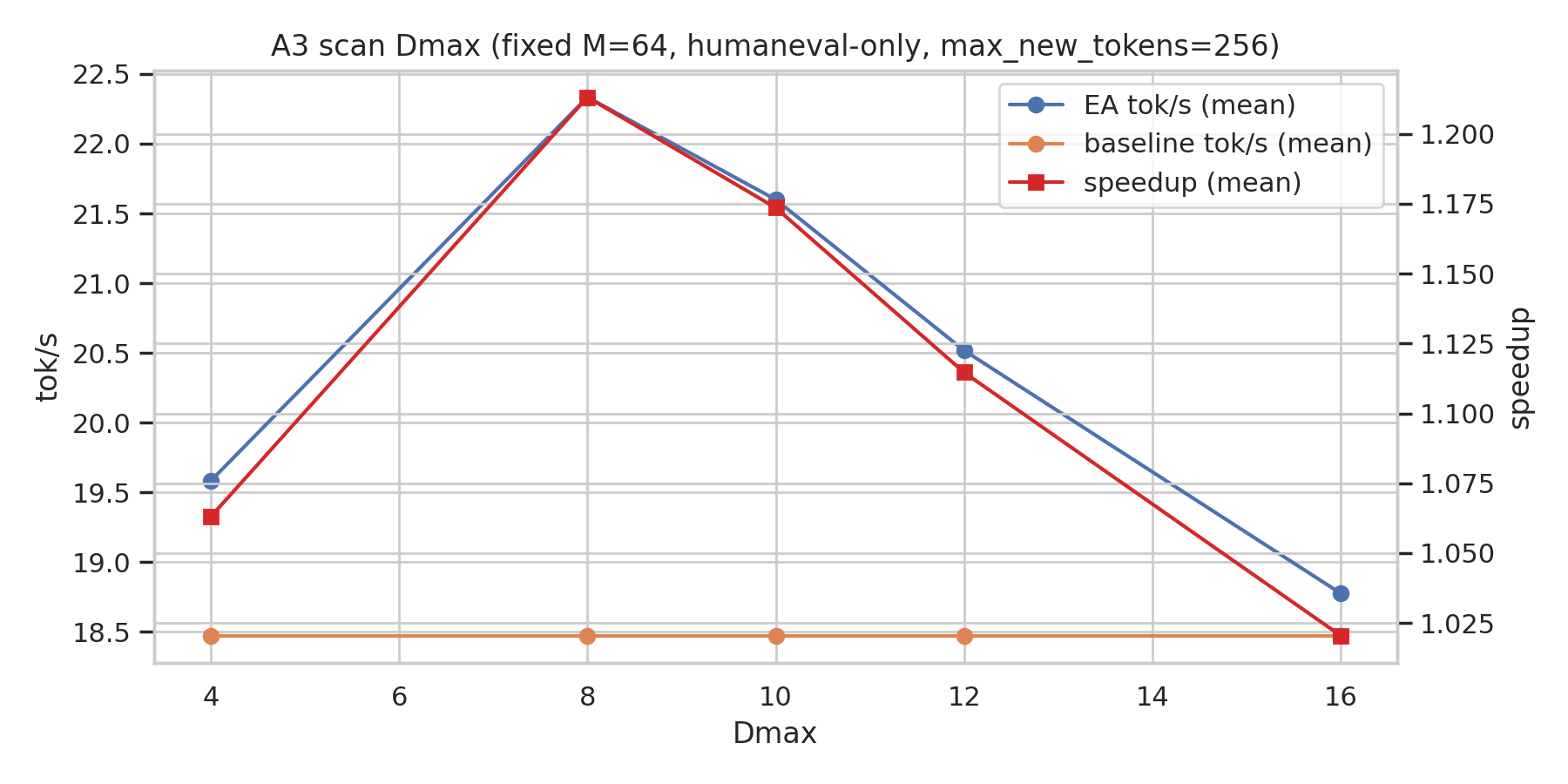}
    \caption*{(b) Scan \(D_{\max}\) (fixed \(M=64\)).}
  \end{minipage}
  \caption{Budget sweeps show non-monotonic throughput behavior and configuration-dependent sweet spots.}
  \label{fig:budget_sweeps}
\end{figure}


\noindent\textbf{Result.}
Throughput is non-monotonic in both \(M\) and \(D_{\max}\) (Table~\ref{tab:budget_sweep}, Figure~\ref{fig:budget_sweeps}). The best mean speedup in our sweep is \(1.48\times\) at \(M=16, D_{\max}=10\). Increasing \(M\) or \(D_{\max}\) beyond this regime can reduce speedups, consistent with (i) higher tensorization/masking/commit overheads and (ii) lower acceptance probabilities at deeper draft positions.

\noindent\textbf{Takeaway.}
Tree speculation has a configuration-dependent sweet spot; lightweight budget sweeps (or adaptive policies) are necessary for stable performance in deployment.

\paragraph{E3: Where does time go? Stage breakdown (instrumented profile).}

\textbf{Goal.} Attribute overheads to identify optimization targets beyond the core teacher forward.
\textbf{Setup.} We run an instrumented profile (same prompt set, \texttt{max\_new\_tokens=1024}). Because instrumentation perturbs runtime, we use these measurements for diagnosis rather than headline throughput.

\begin{figure}[t]
  \centering
  \begin{minipage}{0.49\linewidth}
    \centering
    \includegraphics[width=\linewidth]{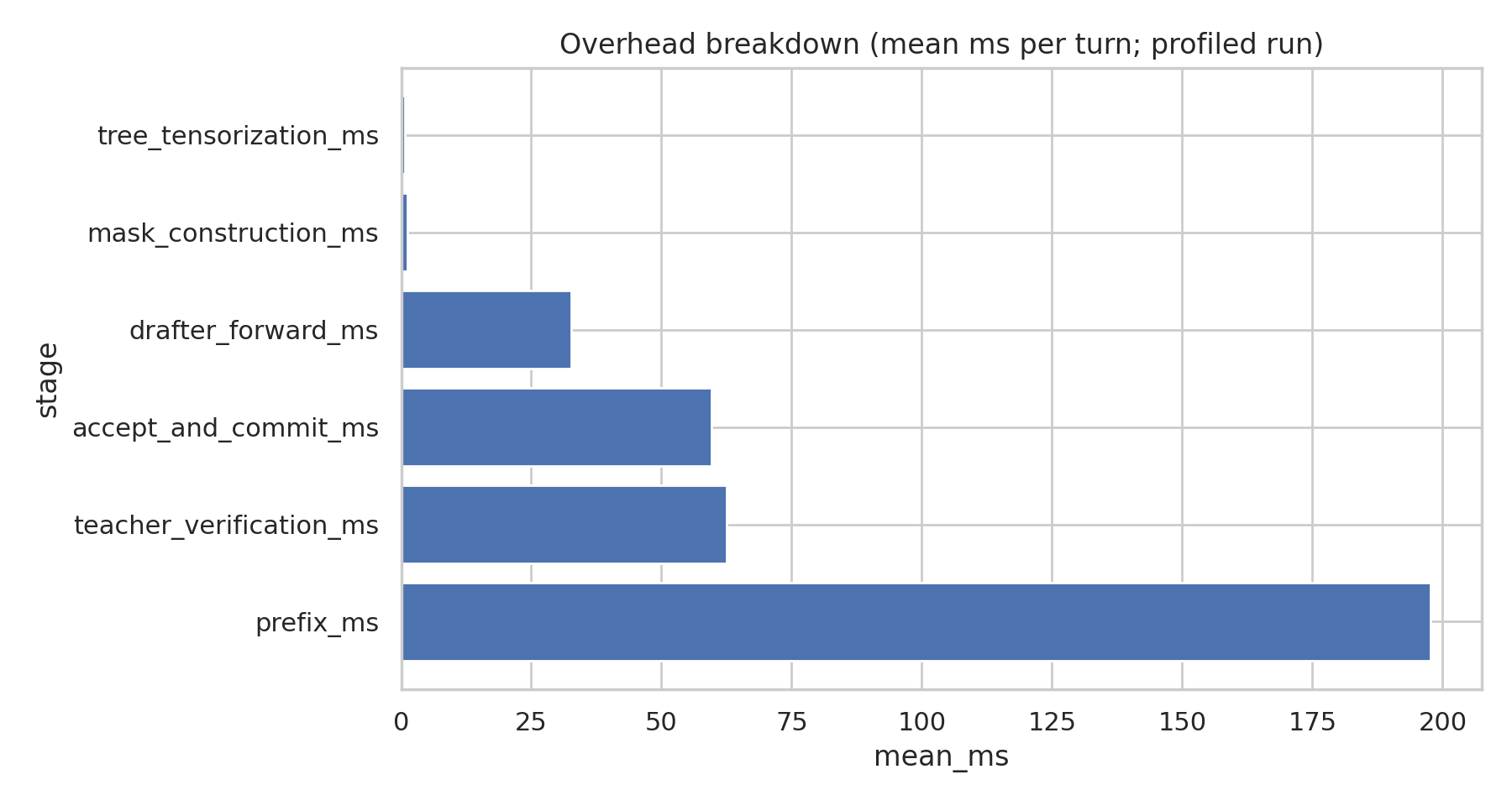}
    \caption*{(a) Mean stage time (ms).}
  \end{minipage}\hfill
  \begin{minipage}{0.49\linewidth}
    \centering
    \includegraphics[width=\linewidth]{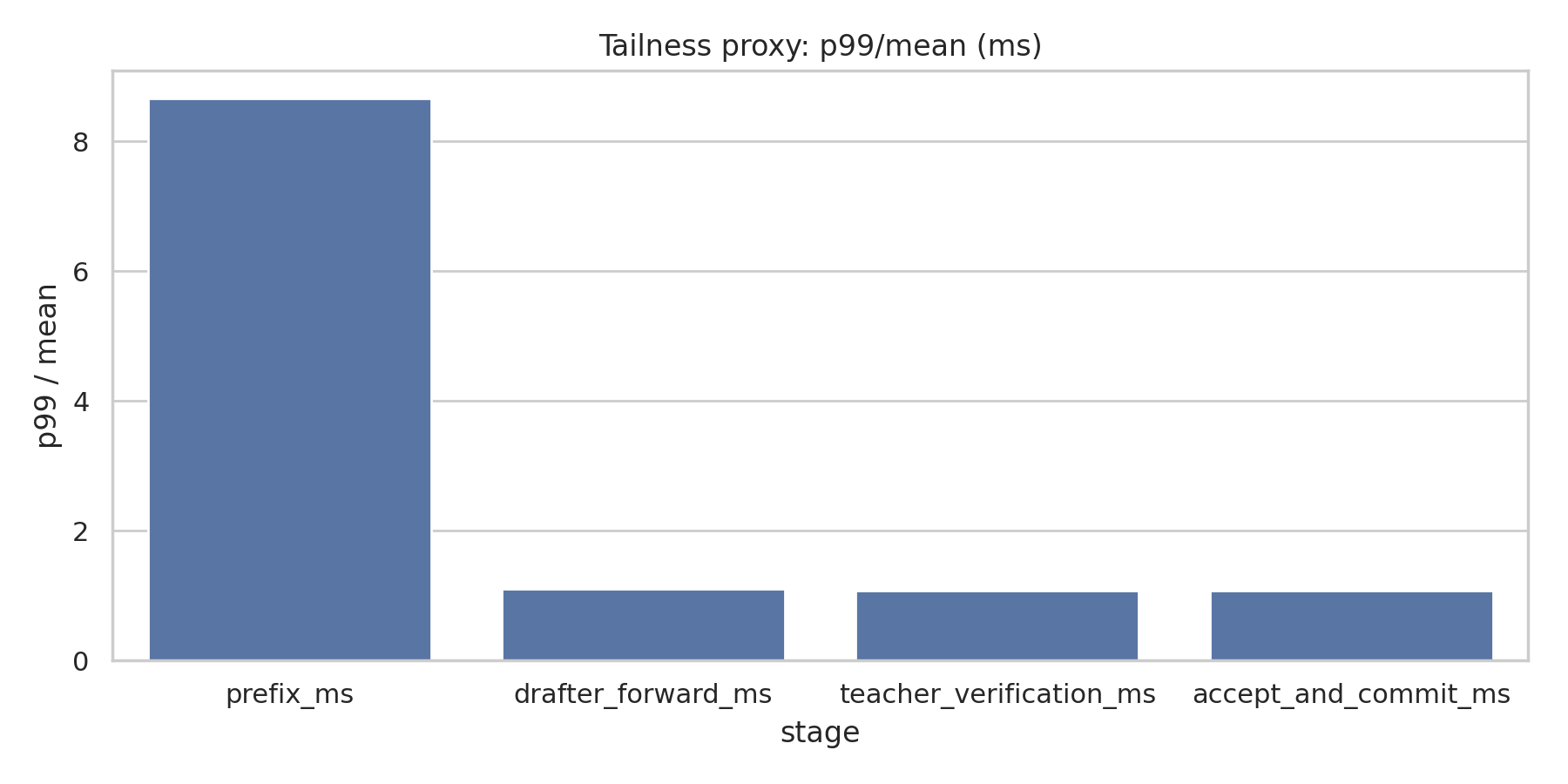}
    \caption*{(b) Tail ratio (p99/mean).}
  \end{minipage}
  \caption{Overhead breakdown (instrumented; analysis-only). Tree tensorization and mask construction are millisecond-scale, while verification and commit are comparable in magnitude. Prefill exhibits a pronounced long tail.}
  \label{fig:overhead_breakdown}
\end{figure}


\noindent\textbf{Result.}
Tree tensorization and mask construction are millisecond-scale on average, while verification and commit are comparable in magnitude. Prefill exhibits a pronounced long tail (Figure~\ref{fig:overhead_breakdown}).

\noindent\textbf{Takeaway.}
Further acceleration should prioritize commit efficiency and long-context prefill behavior; mask construction is not the dominant bottleneck in our stack.

\paragraph{E4: Negative result---fixed-window drafter truncation hurts (and why).}
\textbf{Goal.} Test whether reducing drafter context can improve throughput without harming acceptance.
\textbf{Setup.} We truncate the drafter context to a fixed window \(W\in\{128,256,512\}\) while keeping the teacher context intact, and compare against no truncation.

\begin{table}[t]
\centering
\caption{Drafter-only fixed-window truncation (\texttt{max\_new\_tokens=1024}, fused on). Truncation reduces acceptance and can negate throughput gains.}
\label{tab:truncation}
\begin{tabular}{lrrrr}
\toprule
Window \(W\) & EA Tok/s (mean) & Speedup (mean) & accept\_L mean & accept\_L p90 \\
\midrule
none & 19.93 & 1.15 & 3.17 & 6 \\
128  & 11.97 & 0.69 & 1.48 & 3 \\
256  & 13.22 & 0.76 & 1.68 & 4 \\
512  & 15.14 & 0.87 & 1.99 & 4 \\
\bottomrule
\end{tabular}
\end{table}

\begin{figure}[t]
  \centering
  \includegraphics[width=0.78\linewidth]{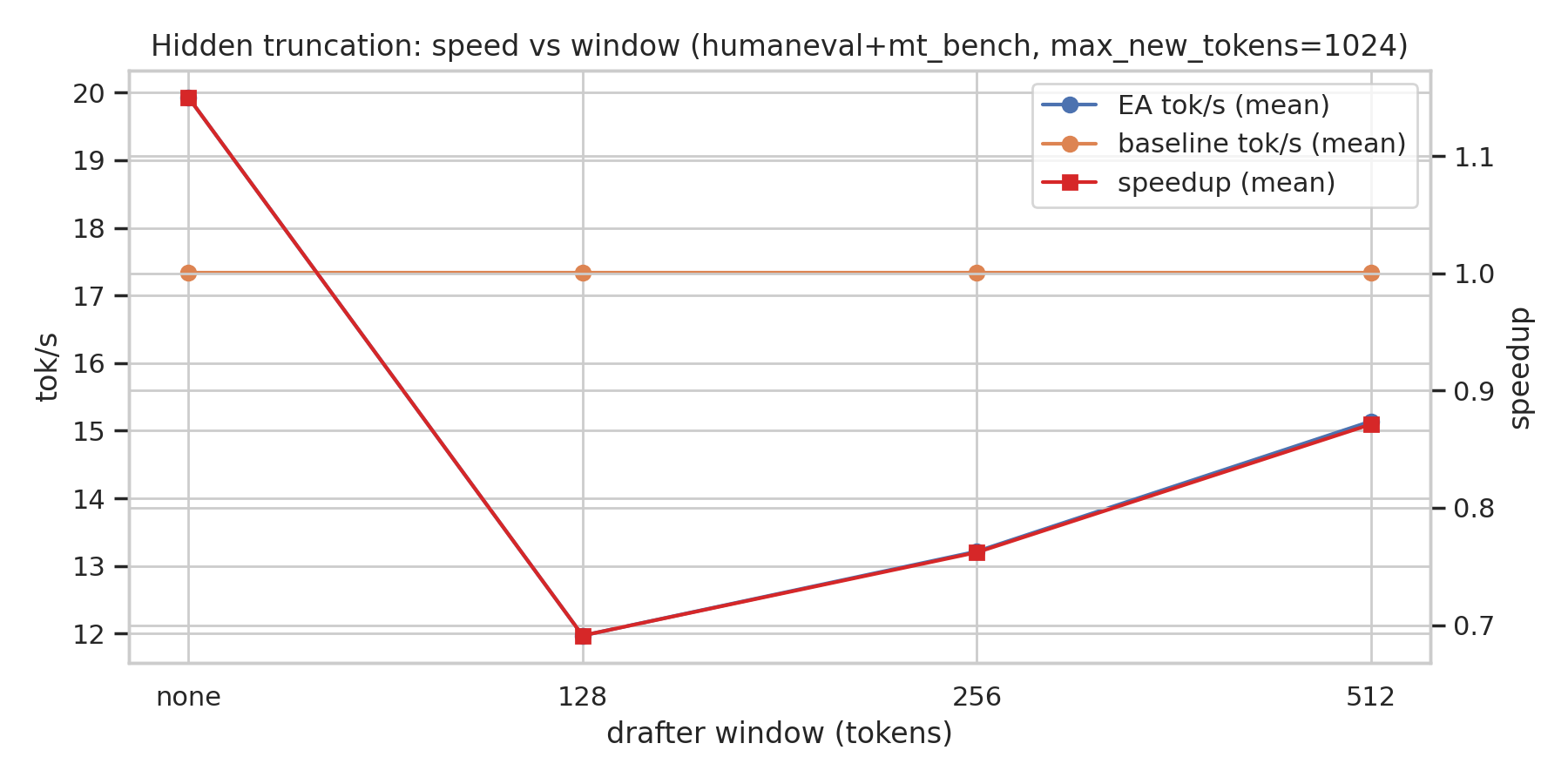}
  \caption{Throughput speedup under drafter-only fixed-window truncation. Smaller windows reduce \(L_k\) and degrade end-to-end speed.}
  \label{fig:trunc_speedup}
\end{figure}

\begin{figure}[t]
  \centering
  \includegraphics[width=0.78\linewidth]{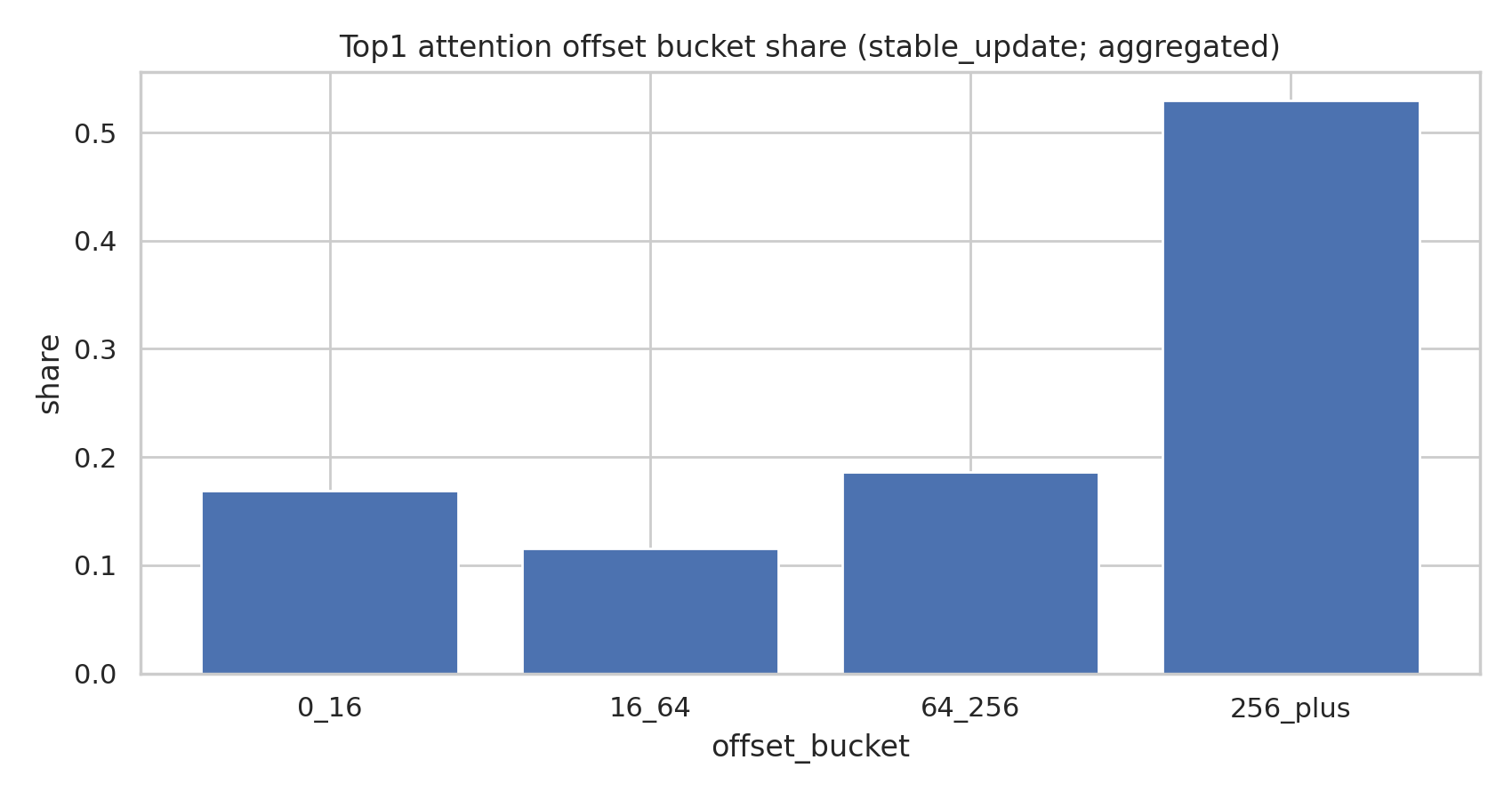}
  \caption{Draft attention evidence (instrumented; analysis-only): the top-1 attention location frequently lies in far history (\(\texttt{256\_plus}\) bucket), consistent with truncation harming draft quality and acceptance.}
  \label{fig:attn_far_history}
\end{figure}


\noindent\textbf{Result.}
Fixed-window truncation substantially reduces acceptance and can reverse speedups (Table~\ref{tab:truncation}, Figure~\ref{fig:trunc_speedup}). Attention profiling provides supporting evidence: the top-1 attention location frequently falls into far-history buckets (Figure~\ref{fig:attn_far_history}), indicating that hard truncation discards positions the drafter often relies on.

\noindent\textbf{Takeaway.}
Naive drafter truncation is counterproductive in this setting; context reduction must be semantics-aware (e.g., adaptive memory, retrieval, or truncation policies informed by attention/importance signals).

\paragraph{Additional analyses (appendix-ready).}
Beyond the above core results, we have structured analyses on how prompt length, synthetic chat history length, and synthetic prefix/context length affect acceptance, tree shape, and per-stage timing.
These results are well-suited for an appendix to strengthen the ``why'' story without expanding the main paper length.

\section{Conclusion}
\label{sec:conclusion}

We presented \textsc{EAGLE-Pangu}, an accelerator-safe port of tree speculative decoding on the Pangu+Ascend stack. Our system combines (i) an explicit branch/commit KV-cache manager, (ii) device-defined tree tensorization that avoids undefined indices, and (iii) a fused-kernel-compatible tree-masked teacher verification path with an eager fallback.

Across our batch-1 benchmark, \textsc{EAGLE-Pangu} achieves substantial throughput gains (up to \(1.27\times\) mean and \(2.46\times\) p99 speedup).

Our experimental design combines lightweight \emph{analysis} studies (mask-leakage checks, acceptance/overhead breakdown, and sensitivity to speculative budget and depth) with \emph{validation} benchmarks that measure end-to-end throughput and latency under both single-request and serving-like batching conditions. Together, these results characterize the practical trade-offs among drafter capacity, speculative budget, batching/padding efficiency, and verification overhead, and demonstrate that tree speculation can reduce teacher invocations and improve decoding efficiency without requiring invasive kernel rewrites.

\paragraph{Limitations and future work.}
First, the speedup is bounded by the drafter cost and by verification overheads (mask construction, cache commit), which become more pronounced as the speculative tree grows; more compact mask representations and deeper kernel fusion are promising directions. Second, performance depends on drafter quality: stronger drafters increase acceptance but may erode gains if drafting becomes compute-bound; speculation-aware distillation and adaptive branching policies could improve this trade-off. Finally, our current evaluation focuses on decoding efficiency; extending the study to longer-context workloads, multi-turn chat with tool use, and multi-device serving will further clarify the deployment envelope.

Overall, this work provides a practical path to deploy tree speculative decoding on Ascend-class accelerators, with explicit correctness invariants and a modular pipeline that can be tuned to different model sizes and serving constraints.

\clearpage
{\small
\bibliographystyle{unsrtnat}
\bibliography{refs}
}

\end{document}